\documentclass[11pt]{article} 
\usepackage{rldmsubmit,palatino}
\usepackage{graphicx}

\usepackage{color}

\usepackage[utf8]{inputenc} 
\usepackage[T1]{fontenc}    
\usepackage{url}            
\usepackage{hyperref}       
\usepackage{booktabs}       
\usepackage{amsfonts}       
\usepackage{nicefrac}       
\usepackage{microtype}      

\usepackage{algorithm, algpseudocode}
\usepackage{mathrsfs}
\usepackage{dsfont}
\usepackage{lmodern}
\usepackage{array}
\usepackage{amsmath}
\usepackage{amssymb}
\usepackage{graphicx}
\usepackage{mathrsfs}
\usepackage{psfrag}
\usepackage{color}
\usepackage{here}
\usepackage{wasysym}
\usepackage{enumitem}
\usepackage{xcolor}
\usepackage{caption}
\usepackage{subcaption}
\usepackage{amsthm}
\usepackage{lipsum}

\newcommand{\Exp}{\mathds{E}}

\DeclareMathOperator*{\argmax}{arg\,max}

\newcommand{\Sc}{\mathcal{S}}
\newcommand{\Ac}{\mathcal{A}}
\newcommand{\Mc}{\mathcal{M}}

\newcommand{\Hc}{\mathcal{H}}
\newcommand{\Oc}{\mathcal{O}}
\newcommand{\Alg}{{\rm alg}}
\newcommand{\E}{\mathbb{E}} 

\title{On Optimistic versus Randomized Exploration \\ in Reinforcement Learning}

\author{
Ian Osband \\
Google Deepmind \\
\texttt{iosband@google.com} \\
\And
Benjamin Van Roy \\
Stanford University \\
\texttt{bvr@stanford.edu} \\
}

\begin{document}

\maketitle

\begin{abstract}
We discuss the relative merits of optimistic and randomized approaches to exploration in reinforcement learning.
Optimistic approaches presented in the literature apply an optimistic boost to the value estimate at each
state-action pair and select actions that are greedy with respect to the resulting optimistic value function.
Randomized approaches sample from among statistically plausible value functions and select actions that
are greedy with respect to the random sample.  Prior computational experience suggests that randomized
approaches can lead to far more statistically efficient learning.  We present two simple analytic examples that
elucidate why this is the case.  In principle, there should be optimistic approaches that fare well relative to
randomized approaches, but that would require intractable computation.
Optimistic approaches that have been proposed in the literature sacrifice statistical efficiency for the sake of computational efficiency.
Randomized approaches, on the other hand, may enable simultaneous statistical and computational efficiency.
\end{abstract}


\acknowledgements{This work was generously supported by a research grant from Boeing, a Marketing Research Award from Adobe, and a Stanford Graduate Fellowship,
courtesy of PACCAR.}

\startmain 

\section{A Reinforcement Learning Problem}
\label{sec: problem formulation}

We consider the problem of learning to optimize a random finite-horizon MDP
{\medmuskip=0mu
\thinmuskip=0mu
\thickmuskip=0mu
$\mathcal{M} = (\Sc, \Ac, \mathcal{R}, \mathcal{P}, H, \rho)$} over episodes of interaction, where
$\Sc =\{1,..,S\}$ is the state space, $\Ac=\{1,..,A\}$ is the action space, $H$ is the horizon, and $\rho$ is the initial state distribution.
At the start of each episode the initial state $s_0$ is drawn from the distribution $\rho$.
In each time period $t=0, \cdots, H-1$ within an episode, the agent observes state $s_t \in \Sc$, selects action $a_t \in \Ac$,
receives a reward $r_{t+1} \sim \mathcal{R}_{t, s,a}$, and transitions to a new state $s_{t+1} \sim \mathcal{P}_{t, s,a}$.
What we consider could be referred to as a Bayesian reinforcement learning setting, in which the unknown episodic nonstationary finite-horizon
MDP $\mathcal{M}$ is taken to be a random variable.

A policy $\pi$ is a mapping from a state $s \in \Sc$ and period $t=0,..,H-1$ to an action $a \in \Ac$.
For each MDP $\mathcal{M} = (\Sc, \Ac, \mathcal{R}, \mathcal{P}, H, \rho)$ and policy $\pi$ we define the state-action value function for each period $t$:
\vspace{-1mm}
\begin{equation}
\label{eq: q value tabular}
  Q^{\mathcal{M}}_{\pi, t}(s, a) := \E_{\Mc,\pi}\left[ \sum_{\tau=t}^{H-1} \overline{r}^{\mathcal{M}}(s_\tau,a_\tau) \Big| s_t = s, a_t=a \right],
\end{equation}
where $\overline{r}_t^{\mathcal{M}}(s,a) = \Exp[ r_{t+1} | \mathcal{M},  s_t=s, a_t=a]$.
The subscript $\pi$ indicates that actions over periods $t,\ldots,H-1$ are selected according to the policy $\pi$.
Let $V^{\mathcal{\Mc}}_{\pi, t}(s) := Q^{\mathcal{M}}_{\pi, t}(s, \pi(s,t))$.
A policy $\pi^{\mathcal{M}}$ is optimal for the MDP $\mathcal{M}$ if $\pi^{\mathcal{M}} \in \argmax_{\pi} V^{\mathcal{M}}_{\pi, t}(s)$ for all $s \in \Sc$ and $t=0,\ldots,H-1$.
We will use $\pi^{\mathcal{M}}$ to denote such an optimal policy.

Let $\mathcal{O}_\ell = (s_0^\ell, a_0^\ell, r_1^\ell, \ldots, s_{H-1}^\ell, a_{H-1}^\ell, r_H^\ell)$ be the sequence of observations made during episode $\ell$.
Let $\Hc_{L-1} = (\mathcal{O}_\ell: \ell=1,\ldots,L-1)$ denote the history of observations made prior to episode $L$.
The agent's behavior is governed by a reinforcement learning algorithm $\Alg$.
Immediately prior to the beginning of episode $L$, the algorithm produces a policy $\pi^L = \Alg(\Sc, \Ac, \Hc_{L-1})$
based on the state and action spaces and the history $\Hc_{L-1} = \left(\Oc_\ell: \ell = 1,\ldots,L-1\right)$ of observations made over previous episodes.
Note that $\Alg$ may be a randomized algorithm, so that multiple applications of $\Alg$ may yield different policies.

In episode $\ell$, the agent enjoys a cumulative reward of $\sum_{t=1}^{H} r^\ell_t$.
We define the {\it regret} over episode $\ell$ to be the difference between optimal expected value and the sum of rewards generated by algorithm $\Alg$.
This can be written as $V_{\pi^\Mc, t}^{\Mc}(s^\ell_0) - \sum_{t=0}^{H-1} r_{t+1}$, where actions are generated by a policy $\pi^\ell$ is produced by algorithm $\Alg$
and state transitions and rewards are generate by MDP $\Mc$.

\section{Optimism versus Randomization}

In principle, given a history $\Hc_{L-1}$ of observations gathered over prior episodes, we can generate a point estimate
$\hat{Q}_t = \E\left[Q^{\Mc}_{\pi^\Mc, t} | \Hc_{L-1}\right]$ of the optimal state-action value function and apply a greedy policy
with respect to this estimate over episode $L$.  However, it is often essential to apply a policy that will explore beyond
this to make discoveries that amplify expected rewards over subsequent episodes.

Optimistic approaches induce exploration by generating optimistic estimates $\overline{Q}_t$
of state-action values and following a greedy policy with respect to optimistic estimates.  The idea is that an
optimistic estimate $\overline{Q}_t(s,a)$ should represent the highest statistically plausible value of $Q^{\Mc}_{\pi^\Mc, t}(s,a)$, given prior
knowledge and observed history.

An alternative approach is to generate prior to each $L$th episode the optimal value function $\tilde{Q}_t$ for an MDP sampled from
the posterior distribution of $\Mc$ conditioned on the history $\Hc_{L-1}$.  This is equivalent to sampling $\tilde{Q}_t$
from the posterior distribution of $Q^{\Mc}_{\pi^\Mc, t}$.  As discussed in \cite{Russo2013b,Osband2013,Russo2014},
such a randomized approach can be analyzed through the study of confidence sets, similarly with how optimistic algorithms
are typically studied, and offer performance similar to well-designed statistically efficient optimistic approaches.  As we will
discuss, the performance advantage of randomized approaches arises from the fact that optimistic
approaches proposed and applied in the literature forgo statistical efficiency for computational tractability.

\begin{figure}[h!]
\centering
  \includegraphics[scale=0.5]{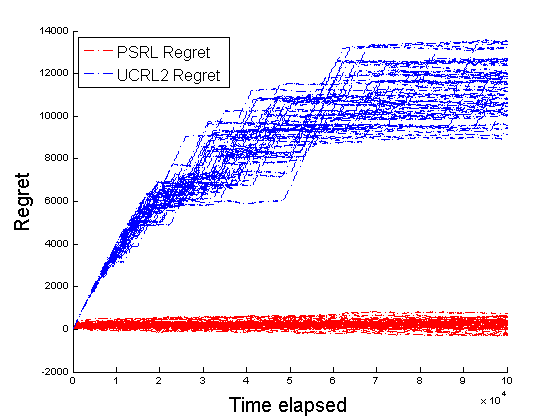}
\caption{Cumulative regret of UCRL2 and PSRL in the \emph{RiverSwim} environment.}
\label{fig:RiverSwim}
\end{figure}

Empirical evidence suggests that randomization often leads to much faster learning than optimiism.
For example, Figure \ref{fig:RiverSwim}, taken from \cite{Osband2013}, plots regret of UCRL2 \cite{Jaksch2010} and PSRL \cite{DeardenFA99,Osband2013}
applied to a variation of the \emph{RiverSwim} problem from \cite{Strehl2006}.  These are well-studied tabular model-based reinforcement learning algorithms that explore via
optimism and randomization, respectively.  For each algorithm, many trajectories are plotted, corresponding
to independent simulations.  For these computations, PSRL began with uninformative Dirichlet priors for transition probabilities
and normal-gamma priors for transition rewards.  It is clear from these results that, for this problem, PSRL learns much faster than UCRL2.
In the next two sections, we present simple analytic examples that provide insight into why randomization offers more desirable
behavior than common optimistic approaches.

\section{Decision Coherence across Time Scales}

\begin{figure}[htpb]
\centering
\includegraphics[scale=0.5]{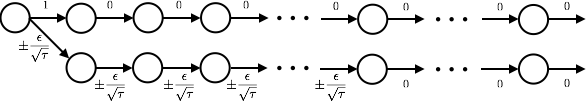}
\caption{Influence of horizon on on exploration decision.}
\label{fig:horizon}
\end{figure}

Consider a simple example illustrated in Figure \ref{fig:horizon}.  An agent is at the left-most state and must select one of two
actions.  Action 1 takes the agent along the ``high road'' over which he knows that he will experience reward of 1 over the first
transition and a reward of $0$ over the following $H-1$ transitions.  Action 2 takes the agent along the low road, where
the agent is uncertain about mean rewards over the first $\tau$ transitions.  According to the agent's posterior distribution, conditioned on the history
of past observations, these mean rewards are independent and identically distributed zero-mean normal random variables
with standard deviation $\epsilon/\sqrt{\tau}$.

Table \ref{tab:horizon} quantifies the posterior distribution of mean value for each action, which is normal with a particular expectation
and standard deviation.  A well-designed optimistic approach should invest to explore if the standard deviation $\epsilon$, which
represents uncertainty in value of the second action, is sufficiently large relative to the difference in expectations, which is $1$.
In particular, the agent should select action $2$ if and only if $c \epsilon > 1$, where $c$ is a tuning parameter that represents the degree
of optimism.  However, ignoring logarithmic factors,
optimistic approaches in the literature (e.g., \cite{Jaksch2010,NIPS2016_6383,DBLP:journals/corr/TangHFSCDSTA16}),
are designed to apply an optimistic boost of the form $c \epsilon \sqrt{\tau}$, which results in selecting action $2$ if and only if $c \epsilon \sqrt{\tau} > 1$.
This is because these optimistic approaches aim to sum over future standard deviations, where
one should more appropriately combine uncertainties by summing variances.  This flaw in uncertainty quantification leads to
an incoherence in decision making: for any fixed $c$, there are time scales $\tau$ for which the agent will explore when it
is not sufficiently uncertain or fail to explore despite sufficient uncertainty.

\begin{table}[htpb]
\centering
\begin{tabular}{|c||c|c|c|}
\hline
action & expected value & standard deviation & optimistic boost \\
\hline
\hline
1 & $1$ & $0$ & $0$ \\
\hline
2 & $0$ & $\epsilon$ & $c \epsilon \sqrt{\tau}$ \\
\hline
\end{tabular}
\caption{Expectation and standard distribution of action value, and a typical optimistic boost, as a function of horizon.}
\label{tab:horizon}
\end{table}

A typical randomized approach would, for this problem, explore in each episode with probability equal to the posterior probability that
the value of action 2 exceeds that of action 1. In particular, randomized approaches allocate effort to exploration proportional
to the chances of gaining actionable information.  This probability does not depend on $\tau$ and therefore does not suffer from the
same sort of incoherence with respect to scalings of $\tau$.

\section{Decision Coherence across Space Scales}

\begin{figure}[htpb]
\centering
    \begin{subfigure}{.25\textwidth}
        \centering
	\includegraphics[scale=0.5]{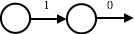}
	\caption{action 1}
	 \label{fig:state1}
    \end{subfigure}
    \begin{subfigure}{.25\textwidth}
        \centering
	\includegraphics[scale=0.5]{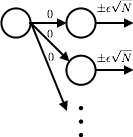}
	\caption{action 2}
	 \label{fig:state2}
    \end{subfigure}
\caption{Influence of number of possible next states on on exploration decision.}
\label{fig:state}
\end{figure}

Now consider an example illustrated in Figure \ref{fig:state}.  The diagrams focus on possible transitions from a single state.  Action 1
generates an immediate reward of $1$, and is known to transition to a state that leads to no subsequent value.  Action $2$ generates no
immediate reward and is known to transition to one of $N$ states, each with probability $1/N$.  From each possible next state,
the agent's posterior distribution models subsequent mean value as an independent zero-mean normal random variable with
standard deviation $\sqrt{N}$.

Table \ref{tab:state} quantifies the posterior distribution of mean value for each action, which is normal with a particular expectation
and standard deviation.  A well-designed optimistic approach should invest to explore if the standard deviation $\epsilon$, which
represents uncertainty in value of the second action, is sufficiently large relative to the difference in expectations, which is $1$.
In particular, the agent should select action $2$ if and only if $c \epsilon > 1$, where $c$ is a tuning parameter that represents the degree
of optimism.  However, ignoring logarithmic factors, common optimistic approaches would apply the average among optimistic boosts
$c \epsilon \sqrt{N}$ associated with possible next states, which results in selecting action $2$ if and only if $c \epsilon \sqrt{N} > 1$.
This is because these optimistic approaches average over standard deviations at possible next states, where
one should more appropriately average variances.  This flaw in uncertainty quantification leads to
an incoherence in decision making: for any fixed $c$, there are values of $N$ for which the agent will explore when it
is not sufficiently uncertain or fail to explore despite sufficient uncertainty.

\begin{table}[htpb]
\centering
\begin{tabular}{|c||c|c|c|}
\hline
action & expected value & standard deviation & optimistic boost \\
\hline
\hline
1 & $1$ & $0$ & $0$ \\
\hline
2 & $0$ & $\epsilon$ & $c \epsilon \sqrt{N}$ \\
\hline
\end{tabular}
\caption{Expectation and standard distribution of action value, and a typical optimistic boost, as a function of the number of possible next states.}
\label{tab:state}
\end{table}

A typical randomized approach would again explore in each episode with probability equal to the posterior probability that
the value of action 2 exceeds that of action 1.  For our example, it is easy to see that this probability
does not depend on $N$ and therefore does not suffer from the same sort of incoherence with respect to scalings of the state space.

\section{Closing Remarks}

Reinforcement learning holds promise to provide the basis for an artificial intelligence
that will manage a wide range of systems and devices to better serve society's needs.
To date, its potential has primarily been assessed through learning in simulated systems,
where data generation is relatively unconstrained and algorithms are typically trained
over tens of millions to trillions of episodes.  Migrating this technology to real systems
where data collection is costly or constrained by the physical context calls for a focus
on statistical efficiency.  An important part of that lies in how agents explore when
learning.  Optimism and randomization offer guiding principles
for efficient exploration.  We have presented a couple analytic examples
that shed light on sources of advantage in the efficiency of randomized approaches,
relative to optimistic approaches that have been presented in the literature.
In principle, it should be possible to design optimistic approaches that combine uncertainties
in a more coherent manner and consequently
perform at least as well as randomized approaches, but such approaches
may be computationally intractable.

A recent area of intense research activity focusses on designing value function learning methods that
efficiently explore intractably large state spaces.   One thread of work develops count-based optimistic exploration schemes that operate with
value function learning \cite{NIPS2016_6383,DBLP:journals/corr/TangHFSCDSTA16}.  Though these approaches
may be effective for a range of problems, they suffer from incoherencies of the kind illustrated in our examples and therefore
are likely to forgo a substantial degree of statistical efficiency.  An alternative is offered by methods that sample
statistically plausible parameterized value functions  \cite{osband2016rlsvi,osband2016deep,osband2017}.

\bibliography{reference}

\end{document}